\documentclass{article}
\usepackage{spconf,amsmath,graphicx}

\usepackage{amsthm,amsmath,amssymb}
\usepackage{mathrsfs}

\usepackage{graphicx}
\usepackage{algpseudocode}
\usepackage{enumitem}
\usepackage{url}
\usepackage[utf8]{inputenc} 
\usepackage[T1]{fontenc}    
\usepackage{url}            
\usepackage{booktabs}       
\usepackage{amsfonts}       
\usepackage{nicefrac}       
\usepackage{microtype}      
\usepackage{xcolor}         
\usepackage{multirow}
\usepackage{algorithm}
\usepackage{amsmath, bm}
\usepackage{subfig}

\title{Toward Auto-evaluation with Confidence-based Category Relation-aware Regression}
\name{ Jiexin Wang\sthanks{Equal contributions.}, Jiahao Chen$^\ast$, Bing Su\sthanks{Corresponding author.}}
\address{Gaoling School of Artificial Intelligence, Renmin University of China, Beijing, China\\}
\begin{document}
%
\maketitle 
\begin{abstract}
Auto-evaluation aims to automatically evaluate a trained model on any test dataset without human annotations.
Most existing methods utilize global statistics of features extracted by the model as the representation of a dataset. 
This ignores the influence of the classification head and loses category-wise confusion information of the model. 
However, ratios of instances assigned to different categories together with their confidence scores reflect how many instances in which categories are difficult for the model to classify, which contain significant indicators for both overall and category-wise performances. 
In this paper, we propose a Confidence-based Category Relation-aware Regression ($C^2R^2$) method. 
$C^2R^2$ divides all instances in a meta-set into different categories according to their confidence scores and extracts the global representation from them. 
For each category, $C^2R^2$ encodes its local confusion relations to other categories into a local representation. 
The overall and category-wise performances are regressed from global and local representations, respectively. Extensive experiments show the effectiveness of our method.
\end{abstract}

  \section{Introduction}
\label{sec:intro}
Evaluating the performance of a trained model on given test datasets is crucial. Usually, human annotations of test instances are required to evaluate the performance of the model. However, in the real world, it is difficult to collect and label new instances for constituting a test set for every scenario. To tackle this problem, auto-evaluation aims at predicting the performance of a trained model on given unlabeled test sets.

\begin{figure}[t]
\begin{minipage}[b]{1.0\linewidth}
  \centering
  \centerline{\includegraphics[width=5.cm]{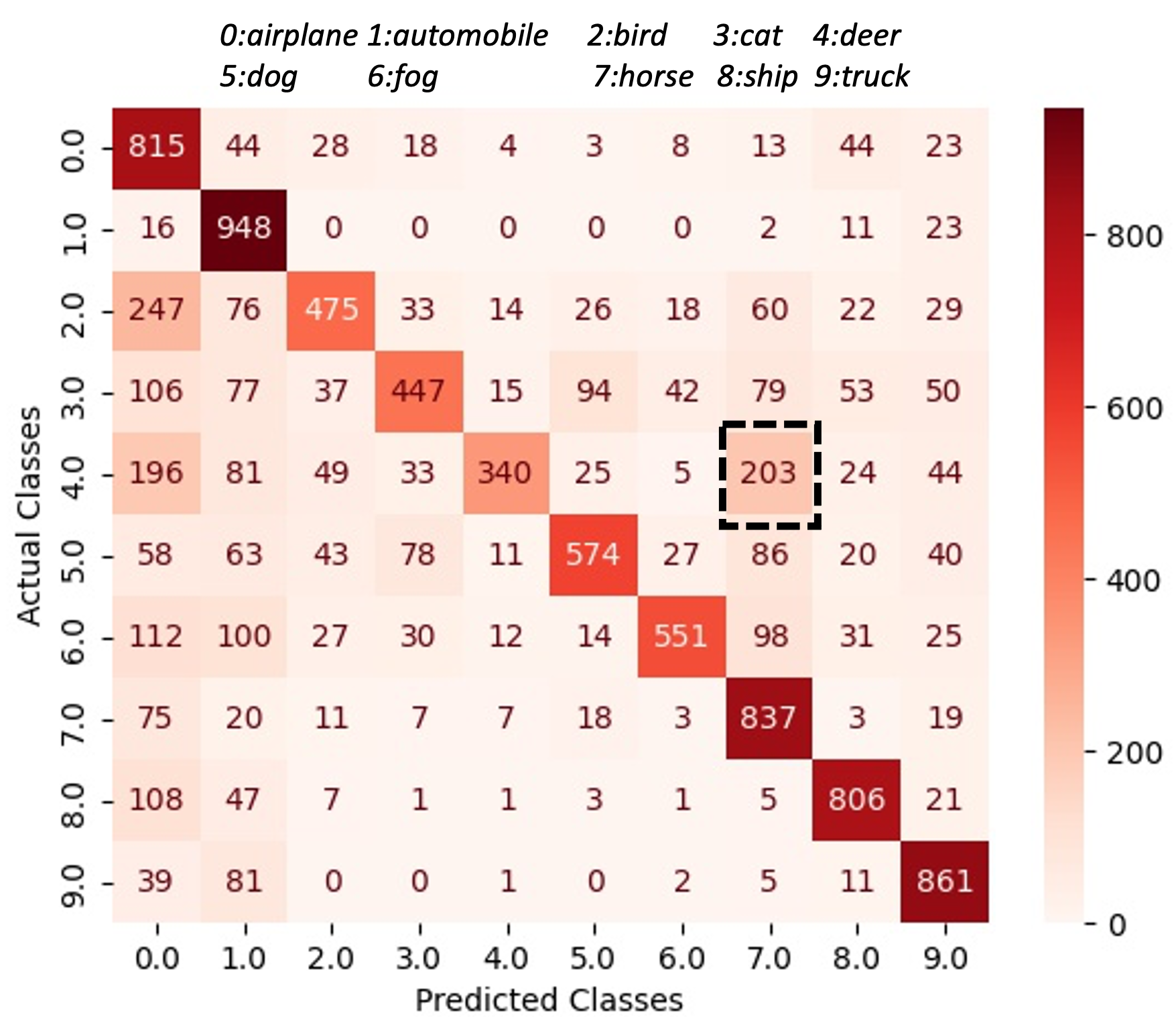}}
\end{minipage}
\caption{The confusion matrix under the CIFAR-10 setup. Instances belonging to "deer" are easily classified into "horse".}
\label{motivation}
\end{figure}
 
\begin{figure*}[t]
  \centering
  \subfloat[Most existing auto-evaluation methods]{ \includegraphics[width=4.6cm]{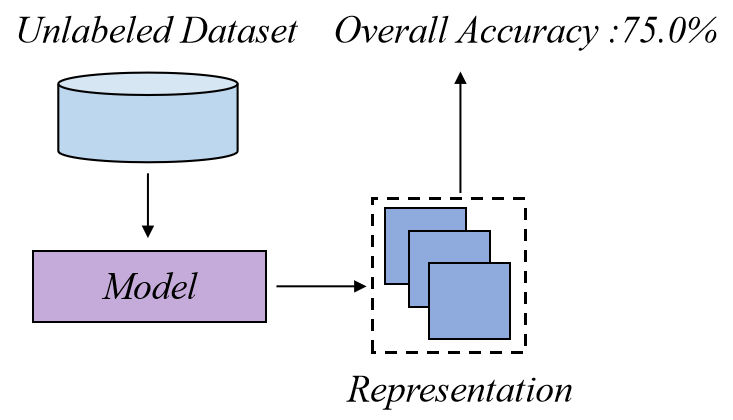}}
  \qquad
  \subfloat[Confidence-based Category Relation-aware Regression method]{\includegraphics[width=8.5cm]{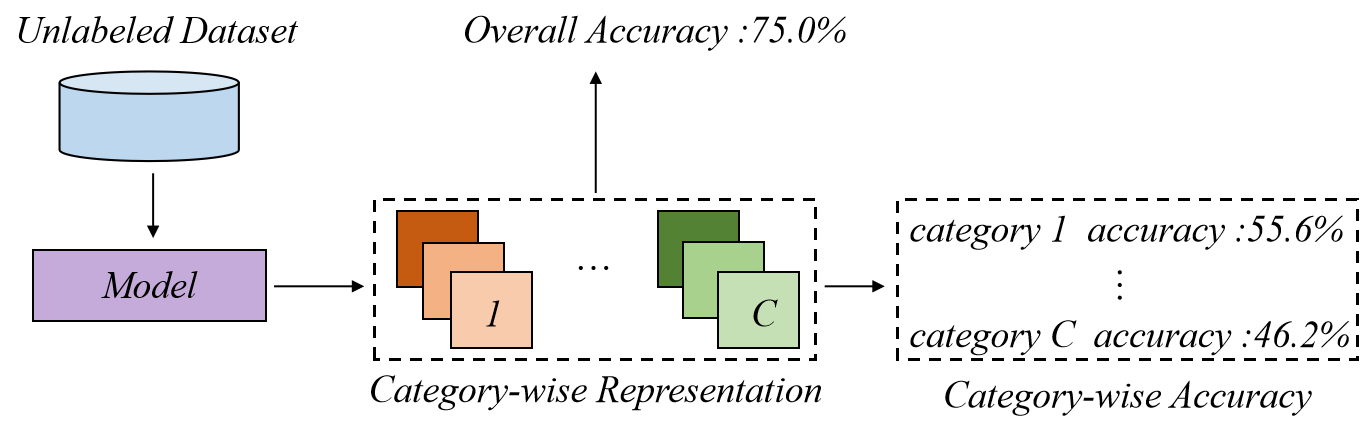}}

\caption{The difference between our method and most existing auto-evaluation methods. (a) Most existing auto-evaluation methods only predict the overall accuracy of the evaluated model on the test set based on the features extracted by the model. (b) Our method can predict the overall accuracy and category accuracies simultaneously by employing relationships among categories.}
\label{methodcompare}
\vspace{-0.1in}
\end{figure*}

Existing methods extract the features of all instances by the model to be evaluated and utilize the statistics of features as the dataset representation. For example, in~\cite{sun2021label}, the distribution shape, cluster center, and representative instances are formed as the semi-structured representation. However, such methods only employ the feature extractor of the model but ignore the influence of its classification head. Given the same feature representation, different classification heads will lead to different performances. A typical solution is using the output of the whole model to construct the representation, which can reflect the confidence of the model that an instance belongs to different categories. Compared with feature-based representations, the confidence score can provide more relevant and direct indications of performance. As shown in Fig.\ref{motivation}, instances belonging to ``deer'' can be misclassified to ``horse'' easily, and this is also reflected in the confidence score that an instance belonging to ``deer'' has relatively comparable confidence to category ``horse''.

Directly replacing features with confidence scores to calculate the statistics as the representation will lose fine-grained evidence in several aspects. 1. Which categories are similar and easily confused? Especially when predicting the accuracy of the model on each specific category, the relationships among categories play a critical role. 
2. How instances from these categories are more difficult to classify? 
High confidence instances are more likely to be classified right while low confidence instances are on the contrary. However, the overall statistics cannot reflect the influence of fine-grained and category-wise confidence scores on the final accuracy.

\begin{figure}[t]
\begin{minipage}[b]{1.0\linewidth}
  \centering
  \centerline{\includegraphics[width=7.cm]{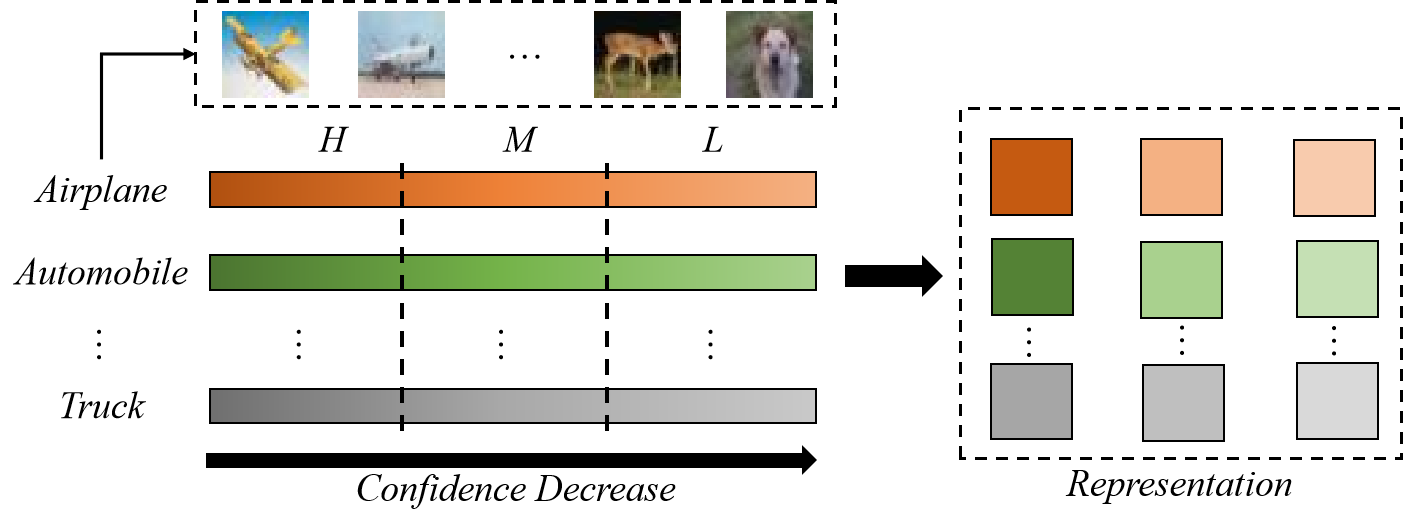}}
\end{minipage}
\caption{The representation of a given dataset extracted by our method. For each category, we collect the instances according to their confidence scores and split the instances into three groups. Statistics of confidence scores in different groups are extracted separately. Statistics from all groups of all categories are combined to form the global representation of this dataset.}
\label{repre}
\vspace{-0.15in}
\end{figure}

Therefore, we propose a new confidence-based category relation-aware ($C^2R^2$) method for the auto-evaluation problem shown in Fig.~\ref{methodcompare}. The main pipeline of $C^2R^2$ is to employ the relationships among categories to extract the representation of a given dataset from instance confidences and utilizes the representation as the input of the regression model. $C^2R^2$ considers the fact that the category-wise confusion-ship conveys strong evidence and different categories provide different contributions to the overall accuracy. As shown in Fig.\ref{motivation}, some categories have low accuracy while others have high accuracy, and different datasets have different difficult categories. We model the influences of different categories explicitly due to their different contributions. Specifically, for each category, we split instances into three groups ($H$, $M$, and $L$) by the confidence of the corresponding category. So instances with similar confidence scores will be gathered together as shown in Fig.\ref{repre}, and each group has its specific information. With the help of three groups, we can extract category information and confidence information comprehensively and combine them as the set representation of a dataset. Moreover, most existing methods only predict the overall accuracy, while our strategy can also realize the prediction of category-wise accuracy. The contribution of this paper are summarized as follows:
\begin{itemize}
    \item [1)] We propose a novel method to encode the fine-grained category-wise relationship information and different levels of confidence scores for different subsets of instances into the hierarchical representation of datasets.
    \item [2)] We propose a novel regression model to utilize our representation. Besides predicting the overall accuracy, it also realizes the prediction of category-wise accuracy not considered in most other auto-evaluation methods. 
    \item [3)] We conduct extensive experiments to evaluate various advanced deep learning models on multiple test sets. The results show the effectiveness of our method.
\end{itemize}

 \section{Related Work}
\textbf{Model generalization.} Understanding the generalization capability of the model is crucial for its application in real-world scenarios~\cite{gulrajani2020search,robey2021model,zhou2021domain}. Generally, due to the mismatch of data distribution between source domain data (training set) and target domain data (test set), the performance of the model in the source domain and target domain is often inconsistent. 
Some attempts have been made to design complexity measures on training models and train sets to predict the generalization gap between train sets and test sets~\cite{corneanu2020computing,liang2017principled,jiang2021assessing,deng2021does}.
Differently, we focus on exploring how to accurately predict the classifier performance by using the distribution difference between the confidence degree of the datasets. 

\textbf{Out-of-distribution (OOD).} Given a trained model, if the test data comes from the same distribution as the training data, it is called in-distribution detection; otherwise, it is called out-of-distribution~\cite{devries2018learning,lee2017training,lin2021mood,zhang2021deep}. There have been many attempts to solve the Auto-evaluation under OOD~\cite{hendrycks2016baseline,liang2017enhancing,jiang2018trust,ji2021predicting}.
These works focus on evaluating the performance of the classification model based on the confidential information of a single sample, without considering the overall information of the test dataset. While our work focuses on how to characterize a dataset from the overall information of the dataset.

\textbf{Set representation.} Set representation has been investigated and widely used in many tasks~\cite{kusner2015word,li2018so,qi2017pointnet,skianis2020rep,naderializadeh2021set}.
Some works trying to solve the Auto-evaluation with the design of set representation~\cite{lee2019set,deng2020labels,sun2021label}.
The above works pay more attention to designing a representation of the overall information in a dataset, while our study also involves designing representations of different categories of information.

 \section{Method}

\subsection{Problem definition}
Given a trained classification model $\mathcal{F}$ and a test dataset $\mathcal{D}^T=\{(\bm{x}_i, y_i)\}_{i=1}^n$, where $y_i \in \{1, \cdots ,C\}$ is the label of $i$-th instance $\bm{x}_i$, $C$ is the number of categories, the normal way to evaluate the final accuracy of the specific model $\mathcal{F}$ on $\mathcal{D}^T$ is to acquire the predication of each instance $\bm{x}_i$ and compare it with the ground-truth label $y_i$. Auto-evaluation aims to automatically evaluate the accuracy of the model $\mathcal{F}$ on any unlabeled test dataset $\mathcal{\widetilde{D}}^T=\{(\bm{x}_i)\}_{i=1}^n$, provided a labeled auxiliary validation set $\mathcal{D}^V$. Formally, we define $Acc(\mathcal{D}^T, \mathcal{F})=\{A_1, \cdots, A_C, A\}$, where $A_1, \cdots, A_C$ denotes the category accuracy and $A$ denotes the overall accuracy of $\mathcal{F}$ on $\mathcal{\widetilde{D}}^T$. The goal is to estimate the value $Acc(\mathcal{\widetilde{D}}^T, \mathcal{F})$. A dominant solution is to generate meta-sets $\mathcal{M}=\{\mathcal{D}^V_1, \cdots, \mathcal{D}^V_k\}$ and constitute meta training sets $\{(\mathcal{H}(\mathcal{D}_i^V, \mathcal{F}), Acc(\mathcal{D}_i^V, \mathcal{F}))\}_{i=1}^k$, where $\mathcal{H}$ denotes the set representation calculator, and datasets $\mathcal{D}^V_i, i=1,\cdots. k$ are obtained by applying different transformations on the labeled validation set $\mathcal{D}^V$ following \cite{deng2020labels,sun2021label}. The meta training sets are then utilized to train a regression model $\mathcal{G}$. After $\mathcal{G}$ is trained, it can be used to predict $Acc(\mathcal{\widetilde{D}}^T, \mathcal{F})$.

\begin{figure}[t]
    \centering
    \includegraphics[width=1\columnwidth]{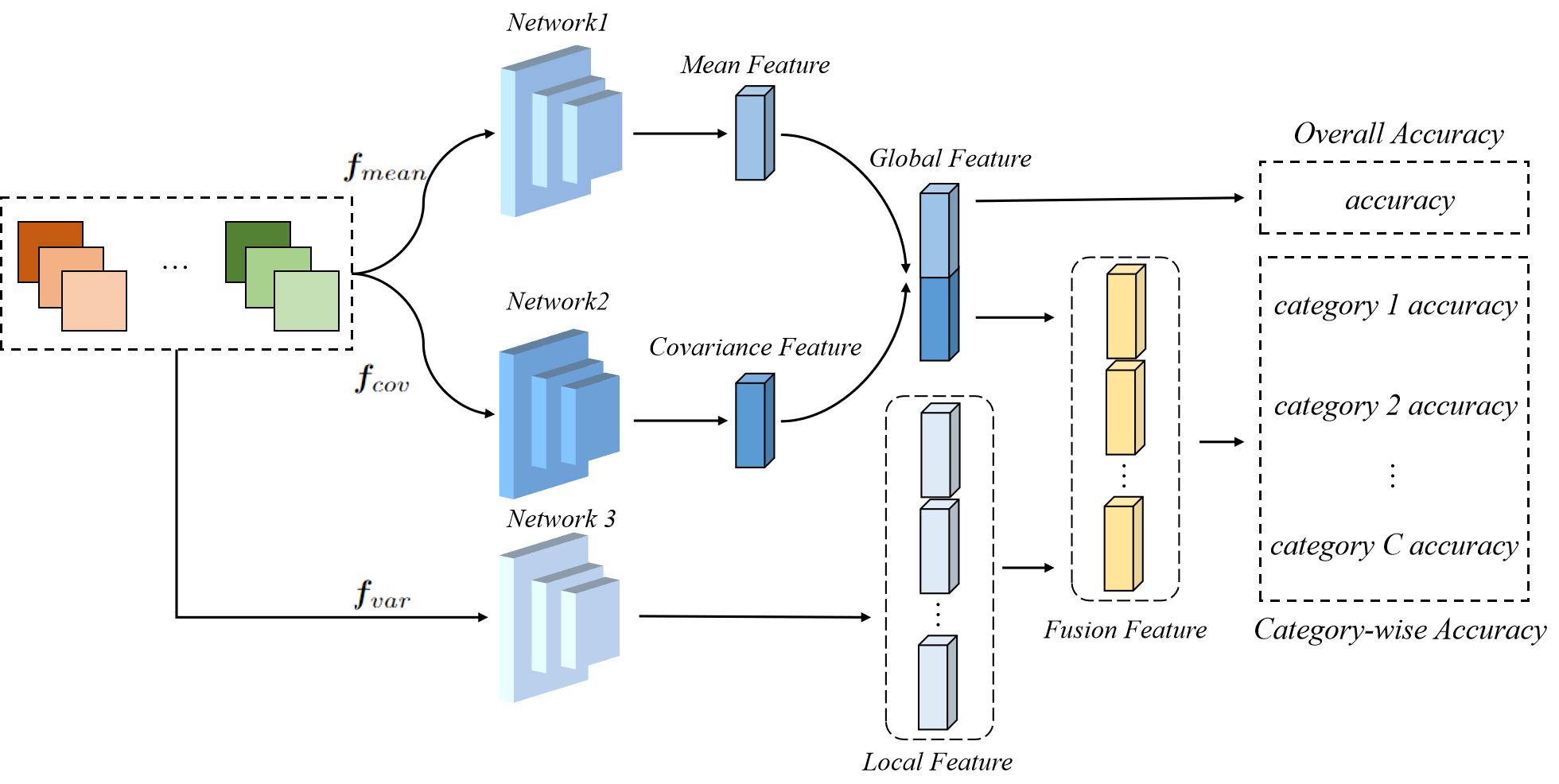}
    \caption{An overview of $C^2R^2$. The overall accuracy is regressed from the global feature. The global feature and the category-wise variance-based local feature are combined as the fusion feature to regress category-wise accuracies.}
    \label{model}
    \vspace{-0.15in}
\end{figure}

\subsection{Set representation}
Due to the high dimensionality, the whole dataset cannot be directly used as the input of $\mathcal{G}$ and we use confidence vectors to constitute the set representation. For 
$\bm{x}_i \in \mathcal{D}$, we obtain its corresponding confidence vector $\bm{z}_i$ and constitute the confidence set $\{\bm{z}_i \in \mathbb{R}^C|\bm{z}_i=softmax(\mathcal{F}(\bm{x}_i)), \bm{x}_i \in \mathcal{D}\}$. The naive solution is to calculate the statistics of the confidence set, which will lose fine-grained evidence, e.g., confidence-based and category-wise information. To capture both information comprehensively, we divide the whole confidence set into category-wise subsets and utilize subset statistics. Specifically, for category $c \in \{1, \cdots , C\}$, we set three groups collecting high confidence instances, medium confidence instances, and low confidence instances, which are denoted as $H_c$, $M_c$, and $L_c$, respectively. Then we extract $\bm{\mu}(\cdot) \in \mathbb{R}^C$, $\bm{\Sigma}(\cdot) \in \mathbb{R}^C$ and $\bm{\sigma}(\cdot) \in \mathbb{R}^C$ for each group, respectively. Taking $H_c$ as an example, these statistics can be calculated as follows: 
\begin{equation}
    \bm{\mu}(H_c) = \frac{1}{|H_c|}\sum_{\bm{z}_i \in H_c}\bm{z}_i
    \label{statistics1}
\end{equation}
\begin{equation}
    \bm{\Sigma}(H_c) = \frac{1}{|H_c|}\sum_{\bm{z}_i \in H_c} (\bm{z}_i - \bm{\mu}(H_c)) \left((\bm{z}_i - \bm{\mu}(H_c))[c] \right)
    \label{statistics2}
\end{equation}
\begin{equation}
    \bm{\sigma}(H_c) = \frac{1}{|H_c|}\sum_{\bm{z}_i \in H_c}(\bm{z}_i - \bm{\mu}(H_c))^2
    \label{statistics3}
\end{equation}
Statistics in $M_c$ and $L_c$ can be extracted in the same way. $\bm{\mu}(\cdot)$ reflects the average confidence vector of a group. The calculation of $\left(\bm{z}_i - \bm{\mu}\left(H_c\right)\right)[c]$ denotes the $c$-th element of $(\bm{z}_i - \bm{\mu}(H_c))$ and $\bm{\Sigma}(\cdot)$ reflects the relationship between category $c$ and other categories. $\bm{\sigma}(\cdot)$ reflects the variation of confidence vectors. Since different groups have different confidence information, we aggregate  $\bm{\mu}(\cdot)$, $\bm{\Sigma}(\cdot)$ and $\bm{\sigma}(\cdot)$ of all groups in the $c$-th class and obtain confidence-based representation $\bm{f}_c^1 \in \mathbb{R}^{3 \times C}$, $\bm{f}_c^2 \in \mathbb{R}^{3 \times C}$ and $\bm{f}_c^3 \in \mathbb{R}^{3 \times C}$, respectively. The representation of category $c$ is $\bm{f}_c=[\bm{f}_c^1, \bm{f}_c^2, \bm{f}_c^3] \in \mathbb{R}^{3 \times 3 \times C}$. For other categories, we follow the same manner to generate category-wise representation and combine them as our set representation $\mathcal{H}(\mathcal{D}, \mathcal{F})=[\bm{f}_1, \cdots, \bm{f}_C]$.

\subsection{Multi-branch Regression model}
For the input of $\mathcal{G}$, we reorganize the shape of the set representation and generate $\bm{f}_{mean} \in \mathbb{R}^{3 \times C \times C}$, $\bm{f}_{cov} \in \mathbb{R}^{3 \times C \times C}$ and $\bm{f}_{var} \in \mathbb{R}^{3 \times C}$. The overall multi-branch architecture is shown in Fig.\ref{model}, we use different networks to further extract features from $\bm{f}_{mean}$ and $\bm{f}_{cov}$ and concatenate the outputs as the global feature to regress the overall accuracy directly. We further add $\bm{f}_{var}$ of the corresponding category to the global feature and generate the fusion feature to regress category-wise accuracy. 
Specifically,  we combine the information of all categories and generate $\bm{f}_{mean}$ and $\bm{f}_{cov}$, respectively, as follows: 
\begin{equation}
    \bm{f}_{mean}= \bm{f}_1^1 \otimes \bm{f}_2^1 \cdots \otimes \bm{f}_C^1
    \label{mean}
\end{equation}
\begin{equation}
    \bm{f}_{cov}= \bm{f}_1^2 \otimes \bm{f}_2^2 \cdots \otimes \bm{f}_C^2
    \label{cov}
\end{equation}
where $\otimes$ denotes the stack operation. For overall accuracy prediction, $\bm{f}_{mean}$ and $\bm{f}_{cov}$ can give confidence-based and category-wise information explicitly. However, it is not sufficient for category accuracy prediction. Therefore, we use $\bm{f}_{var}$ and combine it with the global feature to regress category accuracy. For example, to predict the accuracy of category $c$, $\bm{f}_{var}$ is calculated in Eq.(\ref{var}). 
\begin{equation}
    \bm{f}_{var}= (\bm{f}_1^3[c]) \otimes (\bm{f}_2^3[c]) \cdots \otimes (\bm{f}_C^3[c])
    \label{var}
\end{equation}
We denote the $[c]$ operation as selecting the $c$-th category variance. Also, $\bm{f}_{var}$ considers the relationship of other categories, and $\bm{f}_d^3[c], d \neq c$ gives another view of the variance of category $c$.

We apply the $L_2$ loss to train $\mathcal{G}$ and regress the overall accuracy and category accuracies, respectively. For better convergence of $\mathcal{G}$, the parameters of the main branch (Network1, Network2, and Network3 in Fig.\ref{model}) are only updated by the loss of overall accuracy, and the influence of category accuracy is detached, which can avoid the problem that the model is difficult to converge with the growing number of categories.

 \section{Experiment}
\subsection{Datasets and baselines}
To verify our method, we conduct experiments on image classification tasks (MNIST, CIFAR-10, TinyImageNet). 
Normally, we train a classification model on a train set, generate meta-sets via applying transformations on the valid set, and evaluate the performance of the method on unseen test sets.

\begin{table}[t]
\scriptsize
\setlength{\tabcolsep}{1pt}
\caption{Comparison result of the classifier accuracy predictions. We report RMSE $(\%)$ of the overall predictive accuracy of the dataset to evaluate the performance of predictions.}
\label{table-compare baseline}
	\begin{tabular}{l|ccc|ccc|c}
		\hline
		\multirow{2}{*}{Method} & \multicolumn{3}{c|}{MNIST}          & \multicolumn{3}{c|}{CIFAR-10}     & TinyImageNet \\ \cline{2-8} 
& SVHN & SVHN-C & Digital-S & CIFAR-10.1 & CIFAR-C & CIFAR-F & TinyImageNet-C \\ \hline
PS($\tau_1$=0.8)  & 0.94  &  2.14    & 10.51     & 3.47      & 1.82   & 8.23  & 6.21 \\
PS($\tau_1$=0.9)  & 3.35  &  2.53    & 11.28      & 0.90       & 1.26   & 7.53  & 8.53 \\ \hline
ES($\tau_2$=0.2)      & 3.33  & 2.47     & 11.17      & 1.55       & 1.79   & 7.68  & 8.47 \\
ES($\tau_2$=0.3)      & 2.26  & 2.35     & 10.78      & 6.16      & 2.01   & 8.82  & 6.35 \\ \hline
FD                      & 0.85  & 3.29     & 11.91      & 0.96       & 1.93   & 7.66  & 8.13        \\
AC                      & 0.82  & 1.89     & 11.04      & 0.84      & 1.53   & 7.41  & 6.26        \\
Rotation                & 2.49  & 2.71     & 11.47      & 4.24       & 1.98   & 8.48  & 8.21        \\ \hline
FD+$\sigma+\tau$        & 2.09  & 3.04     & 11.78      & 0.83       & 1.83   & 7.63  & 7.96        \\
Semi         & 0.76  & 1.87     & 9.52       & \textbf{0.74}       & 1.28   & 7.02  & 5.94        \\ \hline
w/o $\bm{f}_{mean}$  & 0.50 & 2.50 & 9.22 & 2.29 & 4.06 & 3.20 & 6.39 \\ 
w/o $\bm{f}_{cov}$   & 9.51 & \textbf{0.70} & 15.79 & 0.89 & 2.60 & 5.19 & 9.41\\
w/o $\bm{f}_{var}$  & 0.72 & 2.50 & 2.73 & 1.80 & 0.69 & 2.84 &3.89\\ 
Ours                    & \textbf{0.36}  & 2.47 & \textbf{2.61} & 1.75       & \textbf{0.58}   & \textbf{2.71} & \textbf{3.63}  \\ \hline 
	\end{tabular}
\vspace{-0.15in}
\end{table}

\begin{table}[t]
\scriptsize%
\centering
\setlength{\tabcolsep}{1pt}
\caption{Result of category prediction performance. We report RMSE $(\%)$ on the category dimension.}
\label{table-cls result}
	\begin{tabular}{l|ccc|ccc|c}
		\hline
		\multirow{2}{*}{Method} & \multicolumn{3}{c|}{MNIST}          & \multicolumn{3}{c|}{CIFAR-10}     & TinyImageNet \\ \cline{2-8} 
		& SVHN & SVHN-C & Digital-S & CIFAR-10.1 & CIFAR-C & CIFAR-F & TinyImageNet-C \\ \hline
		PS($\tau_1$=0.8)  & 23.73 & 24.97 & 17.65 & \textbf{3.34} & 29.95 & 9.39 & 21.97 \\
		PS($\tau_1$=0.9)  & 26.79 & 25.29 & 22.88 & 6.79 & 20.28 & 8.07 & 19.10 \\ \hline
		ES($\tau_2$=0.2)     & 14.75 & 22.97  & 15.69  & 6.74  & 48.73 & 17.18  & \textbf{6.72} \\
		ES($\tau_2$=0.3)     & 20.53 & 17.84  & 26.68  & 10.40 & 63.83 & 23.67 & 10.95 \\ \hline
		AC & 20.07 & 16.95 & 26.56 & 7.12 & 29.16 & 19.38 & 20.89 \\ \hline
 w/o $\bm{f}_{mean}$ & 21.64 & 11.77 & 24.49 & 4.01 & 11.20 & 8.04 &10.96\\ 
w/o $\bm{f}_{cov}$  & 26.07 & \textbf{8.55} & 27.85 & 3.72 & 12.48 &  8.43 &10.87\\
w/o $\bm{f}_{var}$ & 8.99 & 14.07 & 9.75 & 5.80 & 14.37  & 12.62 & 16.07 \\
		Ours & \textbf{7.67} & 14.72 & \textbf{8.09} & 3.68 &  \textbf{12.20} & \textbf{7.88} & 10.53 \\ \hline
	\end{tabular}
 \vspace{-0.15in}
\end{table} 

\subsection{Experiment Results}
In Table.\ref{table-compare baseline}, we report overall accuracy estimation based on various auto-evaluation methods on three groups of unseen datasets. All data, except our proposed method, are from~\cite{sun2021label}. We conduct 10 experiments for our method and report the average results. Our observations are as follows.

\textbf{Our regression model is more suitable to predict classifier performance}.  
In Table.\ref{table-compare baseline}, we observe that the proposed method achieves competitive predictions. On the synthetic unseen test sets (-C), our method's performance is 0.58\%, and 3.6\%, respectively, outperforming the competing methods. On the real-world test sets, the results on Digital-S and CIFAR-F by baselines are all higher than 9.5\% and 7.0\%, respectively, while our model reaches 2.6\% and 2.7\%.
As for our model does not perform well in SVHN-C and CIFAR-10.1, the inner reason is that the meta-sets-based learning task depends on the design of the meta-sets, so the quality of meta-sets has a great impact on this task.
Besides, we conducted additional experiments on the design of the multi-branch network to emphasize its effect, we observe that different components ($\bm{f}_{mean}$, $\bm{f}_{cov}$ and $\bm{f}_{var}$) have different ability to describe a dataset, and our model is more suitable to predict classifier performance.

\textbf{Category-wise prediction performance}. Table.\ref{table-cls result} compares category prediction accuracy errors on category dimension. We observe that: $1)$ score-based methods are sensitive to thresholds, and $2)$ our model achieves promising performance across all setups. We also ablate $\bm{f}_{mean}$, $\bm{f}_{cov}$, and $\bm{f}_{var}$ as a baseline and find that our regression model performs better in most experiments, indicating that these components provide different fine-grained information describing the category.

\begin{table}[t]
\scriptsize
\centering
\setlength{\tabcolsep}{1.5pt}
\caption{Effectiveness of the method designs on the performance. \checkmark denotes that the corresponding confidence feature is applied in the experiment. 
RMSE (\%) is used.}
\label{table-ablation}
\begin{tabular}{ccc|ccc|ccc}
\hline
\multicolumn{3}{c|}{Confidence Feature}  & \multicolumn{3}{c|}{MNIST} & \multicolumn{3}{c}{CIFAR-10}  \\ \cline{1-9} 
 High & Medium & Low &
  \multicolumn{1}{c}{SVHN} &
  \multicolumn{1}{c}{SVHN-C} &
  \multicolumn{1}{c|}{Digital-S} &
  \multicolumn{1}{c}{CIFAR-10.1} &
  \multicolumn{1}{c}{CIFAR-C} &
  \multicolumn{1}{c}{CIFAR-F} \\ \hline
\checkmark & &  &  1.06 &  3.81 & 4.12 & 1.30  & 1.83 & 4.07 \\
& \checkmark &  & 0.75 & 7.87 & 7.57 & 0.86 & 11.20 & 6.47  \\
& & \checkmark & 1.63 & 12.04 & 5.11 & 2.20 & 25.99  & 8.70\\
& \checkmark & \checkmark & \textbf{0.25} & 9.21 & 4.49 & \textbf{0.33} & 7.97 & 4.46  \\ 
\checkmark & & \checkmark & 0.89 & 2.60 & 2.95 & 1.68 & 1.03 & 3.18\\ 
\checkmark & \checkmark & & 0.29 & 2.76 & 2.63 & 1.22 & 2.32 & 3.12\\ \hline
\checkmark & \checkmark & \checkmark & 0.36 & \textbf{2.47} & \textbf{2.61} & 1.75 & \textbf{0.58}  & \textbf{2.71} \\ \hline
\end{tabular}
\vspace{-0.15in}
\end{table}

\textbf{Ablation study: effectiveness of the method designs on the performance}.  
Table.\ref{table-ablation} indicates that the confidential information of each group has varying abilities to represent different datasets, with the best performance achieved when all three groups are applied in most cases. Although SVHN and CIFAR-10.1 achieve the best performance under the setting dropping the high confidence information, adding high confidence information will not lose much performance. Moreover, such information improves performance on other test sets.
Therefore, applying information from all groups is more robust and feasible to characterize datasets and is a better choice.
 \section{Conclusion}
In this paper, we propose a novel method to tackle the auto-evaluation problem by combining the confidence and category information to form a new set representation, and using a multi-branch regression model to predict performance. In addition, we realize the prediction of category accuracy which is ignored in other auto-evaluation methods. Extensive experiments show that our method achieves competitive results compared to various alternative approaches. In the future, we will extend our strategy to other tasks like object detection and segmentation.

\section{acknowledgement}
This work was supported in part by the National Natural Science Foundation of China No. 61976206 and No. 61832017, Beijing Outstanding Young Scientist Program NO. BJJWZYJH012019100020098, and Public Computing Cloud, Renmin University of China.

 \bibliographystyle{IEEEbib}
 \bibliography{refs}
\end{document}